%% file: acl2023.tex
\pdfoutput=1

\PassOptionsToPackage{table,xcdraw}{xcolor}

\documentclass[11pt]{article} 
\usepackage[]{ACL2023}
\usepackage{times}
\usepackage{latexsym}
\usepackage[utf8]{inputenc}
\usepackage{microtype}
\usepackage{inconsolata}
\usepackage[T1]{fontenc}
\usepackage{changepage}
\usepackage{algpseudocode}

\input{math_commands.tex}

\usepackage{graphicx}
\usepackage{booktabs,subcaption,amsfonts,dcolumn}
\usepackage{multirow}
\usepackage[table,xcdraw]{xcolor}
\usepackage{hyperref}
\usepackage{url}
\usepackage{cleveref}
\usepackage{comment}
\crefname{equation}{Eq.}{Eq.}
\crefname{section}{Section}{Sections}
\crefname{appendix}{Appendix}{Appendix}
\crefname{subsection}{Section}{Sections}
\crefname{subsubsection}{Section}{Sections}
\crefname{figure}{Figure}{Figures}
\crefname{table}{Table}{Tables}
\crefname{subfigure}{Figure}{Figures}
\crefname{algocf}{Algorithm}{Algorithms}

\title{Synthetic Pre-Training Tasks for Neural  Machine Translation}

\author{Zexue He\textsuperscript{\rm 1}$^*$,
Graeme Blackwood\textsuperscript{\rm 2}$^*$,  Rameswar Panda\textsuperscript{\rm 2}, \\ \textbf{Julian McAuley\textsuperscript{\rm 1}, Rogerio Feris\textsuperscript{\rm 2}}\\
\textsuperscript{\rm 1}University of California, San Diego \\
\textsuperscript{\rm 2}MIT-IBM Watson AI Lab, IBM Research\\
\textsuperscript{\rm 1}\texttt{\{zehe,jmcauley\}@ucsd.edu} \\
\textsuperscript{\rm 2}\texttt{\{blackwood,rpanda,rsferis\}@us.ibm.com}
}

\begin{document}

\maketitle

\begingroup\def\thefootnote{*}\footnotetext{Equal contribution}\endgroup

\begin{abstract}
Pre-training models with large crawled corpora can lead to issues such as toxicity and bias, as well as copyright and privacy concerns. 
A promising way of alleviating such concerns is to conduct pre-training with synthetic tasks and data, since no real-world information is ingested by the model.
Our goal in this paper is to understand the factors that contribute to the effectiveness of pre-training models when using synthetic resources, particularly in the context of neural machine translation. We propose several novel approaches to pre-training translation models that involve different levels of lexical and structural knowledge, including: 1) generating obfuscated data from a large parallel corpus 2) concatenating phrase pairs extracted from a small word-aligned corpus, and 3) generating synthetic parallel data without real human language corpora. Our experiments on multiple language pairs reveal that pre-training benefits can be realized even with high levels of obfuscation or purely synthetic parallel data. 
We hope the findings from our comprehensive empirical analysis will shed light on understanding what matters for NMT pre-training, as well as pave the way for the development of more efficient and less toxic models.

\end{abstract}


\section{Introduction and Motivation}
\label{sec:introduction}

Neural Machine Translation (NMT) models depend on large quantities of aligned training data \citep{aharoni-etal-2019-massively,fan2021beyond,nllb-paper}. For many language pairs of interest, however, high quality parallel data is either unavailable or exists only in  limited quantities. Training robust NMT systems with such limited data can be a significant challenge.

Even for high-resource language pairs, parallel data can be noisy and frequently contains toxic speech or biased language. Such problems are particularly acute for comparable corpora crawled automatically from the web \citep{kreutzer-crawled-corpora-quality} since it can cause catastrophic mistranslations \cite{costa2022no} or hallucinated toxicity. It is preferable to avoid exposing the model to such data in order to prevent accidental generation of offensive content or egregiously embarrassing translations. Crawled data can also present problematic copyright, attribution, and privacy issues. As an example, the JW300 corpus of Jehovah’s Witnesses publications \citep{agic-vulic-2019-jw300} was recently withdrawn due to a copyright infringement claim.

\begin{figure}[t]
\centering
\includegraphics[width=0.9\linewidth]{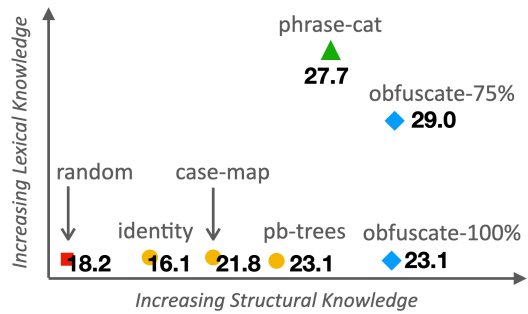}
\vspace{-1.5em}
\caption{A comparison of the extent to which the synthetic data generation methods described in Section~\ref{sec:synthetic_pretraining} encode lexical and/or structural translation knowledge. The vertical  axis compares methods with respect to lexical knowledge. The horizontal axis compares structural knowledge. BLEU scores correspond to the Indonesian-to-English translation task described in Section \ref{sec: experiment}.}\vspace{-1em}
\label{fig:three-approaches}
\end{figure}

Our primary motivation is to investigate how knowledge transfer from NMT pre-training can help to avoid or minimize the data issues described above. We study the impact of pre-training and transfer learning on translation tasks by comparing various procedural approaches to synthetic data generation. Each approach has varying degrees of inherited or artificially constructed lexical and structural translation knowledge. The degree to which each method encodes lexical and/or structural translation knowledge is plotted in abstract form in Figure~\ref{fig:three-approaches}. We describe each of our synthetic data generation methods in Section~\ref{sec:synthetic_pretraining}.

Our first approach (\S\ref{sec:obfuscated-pre-training}) studies the extent to which the transfer benefits of regular  pre-training can be realized when using obfuscated or encrypted data. Our obfuscated corpus is derived from real parallel data by mapping the original words to a vocabulary of `nonsense' tokens. Experiments on six different language pairs show that obfuscated pre-training is able to capture much of the transferable knowledge: pre-training with an obfuscation ratio as high as 75\% is still able to achieve BLEU scores close to those obtained by a model pre-trained on the original un-obfuscated parallel data. 

Our second approach (\S\ref{sec:phrase-concatenation}) seeks to maximize the benefit that can be derived from a specific limited quantity of fine-tuning data. We do this by pre-training on newly constructed artificial sentence pairs synthesized directly from the fine-tuning corpus. The synthetic sentence pairs are created by concatenating randomly sampled aligned phrase pairs extracted from the fine-tuning corpus. Although the sentence-level fluency and grammaticality of sentences constructed using this technique are both quite poor, they do retain word- and phrase-level correspondences and local reordering information that can greatly improve translation quality and robustness compared to models trained using only the original fine-tuning data.

Our third approach (\S\ref{section:synthetic-pretraining-tasks}) explores the pre-training impact of important translation phenomena such as alignments and reordering. We pre-train models on procedurally generated synthetic parallel data that does not derive from any real human language corpus. We design three simple synthetic sequence-to-sequence translation tasks and associated data sets. Since our data is procedurally generated, problems of toxicity, attribution and copyright can be avoided. We evaluate the effectiveness of pre-training and transfer for our synthetic tasks in the context of low-resource NMT. Our results show that -- to a surprising degree -- the transfer benefits of pre-training can be realized even with purely synthetic tasks and data. Our analysis shows that structure, in the form of aligned sub-trees, matters in synthetic pre-training for NMT. 

We empirically evaluate the impact of each of our proposed synthetic pre-training methods in low-resource MT settings (\S\ref{sec: experiment}), followed by a discussion and analysis explaining our insights into what makes for a good pre-trained model (\S\ref{sec: analysis_and_discussion}). We also consider the question of model toxicity. We measure the extent of hallucinated toxicity in each synthetic data generation method, showing that synthetic methods can result in substantially reduced toxicity compared to models pre-trained on real parallel corpora.

The primary {\bf contributions} of our paper are as follows: 
(i) we propose several novel synthetic pre-training tasks, that encode varying degrees of structural and lexical knowledge, in order to gain insights into what makes for a good pre-trained NMT model; 
(ii) we conduct a comprehensive empirical evaluation of knowledge transfer in NMT from synthetic data pre-training, considering metrics of both translation quality and toxicity; and 
(iii) we demonstrate that synthetic data is a promising stepping stone towards relieving the data burden in low-resource translation and building more accurate and trustworthy NMT systems.


\section{Related Work}
\label{sec:related-work}

Transferring knowledge from pre-trained language models \citep{devlin-bert-paper,raffel-t5-paper,brown-gpt-paper} is a common technique for ensuring robust NLP downstream task performance. Early work by \citet{zoph-etal-2016-transfer} explored transfer learning for NMT from a model pre-trained on a single language pair. More recently, methods that transfer from large-scale multilingual pre-trained models \citep{conneau-xlmr-paper,liu-mbart-paper,goyal2022flores,nllb-paper} have achieved improved translation performance across a wide range of language pairs. \citet{aji2020neural} conducted a study on pre-training and transfer for low-resource NMT. These works depend on real human language for pre-training and therefore inherit data issues such as toxicity and bias. In contrast, our work studies NMT pre-training and transfer from synthetic data based on `nonsense' words.


Only a few methods have addressed the problem of pre-training from synthetic data in NLP. \citet{krishna2021does} proposed pre-training for summarization using synthetic article and summary pairs derived from manually curated tasks and a vocabulary of nonsense symbols. \citet{sinha2021masked} have shown that masked language model pre-training with limited word-order information can be almost as effective as regular pre-training. \citet{chiang2020pre,chiang-artificial-datasets} show that non-human language data and artificial datasets (e.g.\ nested sequences of parentheses), can still demonstrate knowledge transfer to downstream NLP tasks. \citet{wu2022insights} compare the effect of pre-training on many simple synthetic tasks. 
Our work in this paper represents the first empirical evaluation of synthetic pre-training for neural machine translation. To the best of our knowledge, our proposed synthetic tasks have not been explored in previous work.

The quality of a pre-trained model should not be measured purely by performance. We should also consider trustworthiness \cite{he-etal-2022-controlling, xu2022leashing, he-etal-2021-detect-perturb}. Recent works have noted that translation systems pre-trained on web-scale corpora are prone to produce toxic \citep{costa2022no} or biased outputs \citep{prates2020assessing, cho2021towards, costa2020gender}, and/or present privacy issues \citep{prates2020assessing, kamocki2016privacy}, which reduces user trustworthiness. Bias mitigation for NMT has been well-investigated while privacy and toxicity issues for translation are still not extensively explored \citep{costa2022no}. \citet{wang2021modeling} propose federated neural machine translation to protect privacy such as commercial leakage or copyright. \cite{costa2022no} mitigate toxicity by filtering training data that matches pre-defined multilingual toxic word lists.



\section{Synthetic Pre-Training for NMT}
\label{sec:synthetic_pretraining}

Pre-training followed by fine-tuning is a common approach to training robust NMT models \citep{conneau-xlmr-paper,liu-mbart-paper}. Our motivation is to understand the extent to which the transfer benefits of pre-training can be replicated using synthetic tasks and data. 
In this section, we describe three approaches to the programmatic generation of synthetic data: (i) pre-training with obfuscated parallel data that implicitly preserves certain language properties such as distributional frequencies, (ii) pre-training with synthetic data created by concatenating aligned phrases, and (iii) pre-training with synthetic tasks designed to encourage transfer learning of important translation properties such as long-distance reordering.

\subsection{Pre-Training on Obfuscated Parallel Data}
\label{sec:obfuscated-pre-training}

In order to gain insight into what makes a good pre-trained model, we design an obfuscated pre-training experiment in which the model learns to translate obfuscated source sequences to obfuscated target sequences. The synthetic training data for this experiment is created by obfuscating words in the original parallel data. We define separate 1-to-1 nonsense token vocabulary mappings for the set of all words that occur in the source and target sides of the data: each source word $s_i$ and target word $t_j$ has a corresponding obfuscated nonsense source token $\mathcal{O}_{s_i}$ and target token $\mathcal{O}_{t_j}$. The synthetic pre-training corpus is created by replacing, with probability $R$, each source and target word with its corresponding obfuscated nonsense token. $R$ thus determines the proportion of obfuscated tokens, allowing us to evaluate the extent to which pre-training knowledge transfer occurs with different obfuscation ratios. This method of obfuscation can be viewed as a trivial form of encrypted training. Although the original word identities are obscured, a great deal of useful information such as distributional frequencies, word order, dependency relations, alignments, and grammatical structure remain implicit in the obfuscated data. An example German-English parallel sentence pair and obfuscations at $R=0.25$ and $R=1.00$ (i.e.\ all tokens obfuscated) are shown below:

\begin{table}[!h]
\centering
\begin{tiny}
\hspace{-1em}
\begin{tabular}{l|r|l}
\multirow{2}{*}{$R=0.00$} & src & {\tt Meine zweite Bemerkung ist etwas ernsthafter.} \\ 
& trg & {\tt My second comment is rather more serious.} \\ \cmidrule(lr){1-3}
\multirow{2}{*}{$R=0.25$} & src & {\tt wfnzc zweite Bemerkung ist etwas ernsthafter .} \\ 
& trg & {\tt My IJODB comment is AHBNB more serious .} \\ \cmidrule(lr){1-3}
\multirow{2}{*}{$R=1.00$} & src & {\tt wfnzc kqknd gmlfd tlieb  ghzwa jdfnd engwd} \\ 
& trg & {\tt UKVFB IJODB XRWOB SZEIA AHBNB LATAA MCSDA ETFJA} \\
\end{tabular}
\end{tiny}
\end{table}

\subsection{Pre-Training on Concatenated Phrases}
\label{sec:phrase-concatenation}

In this section, we propose pre-training an NMT model with synthetic parallel data formed by concatenating aligned phrases. The main advantage of aligned phrases is that they are extracted from real parallel data and thus encode both lexical and structural translation knowledge. Lexical knowledge is defined by the word- and phrase-level correspondences between the source and target language. Structural knowledge, encoded by local reordering within aligned phrases, can also be leveraged.

We first extract a collection of aligned phrases $\mathcal{P}$ using the standard recipe implemented in the Moses SMT Toolkit \citep{koehn2007moses}. The accuracy of the aligned phrases depends on the size and quality of the parallel data: we target low-resource MT and assume there is only a limited quantity of parallel data available. We generate synthetic parallel sentence pairs by first sampling a normally distributed phrase length $P$. We sample each phrase position $p=1 \ldots P$ uniformly at random from $\mathcal{P}$. The source and target sentences thus consist of concatenated source and target phrases. The word order within each sampled phrase is fluent and local reordering may also be captured. The boundaries between phrases, however, typically do not respect natural word order or grammar. In spite of these limitations, we show in Section~\ref{sec:phrase-concatenation-experiments} that this simple method of data augmentation can significantly improve the quality of an NMT model when training data is limited. An example Indonesian-to-English synthetic sentence pair, with phrase boundaries indicated by parentheses, is shown below:

\begin{table}[!h]
\centering
\begin{footnotesize}
\begin{tabular}{r|l}
\multirow{2}{*}{src} & {\tt [sejak Wright] [sambil seringkali] [kami]} \\
 & {\tt \hspace{1em} [50 juta mengingat]} \\
\multirow{2}{*}{trg} & {\tt [from Wright] [in most times] [we]} \\
 & {\tt \hspace{1em} [50 millions as]}\\
\end{tabular}
\end{footnotesize}
\end{table}


\subsection{Pre-Training on Synthetic Tasks and Data}
\label{section:synthetic-pretraining-tasks}

In this section, we define three completely synthetic task variants that can be used for NMT pre-training: (1) the identity operation, (2) case-mapping, and (3) permuted binary trees.  All three tasks are based on a procedural data generation model and can thus be used to generate arbitrary quantities of synthetic data. Procedural generation of synthetic parallel sentence pairs allows for complete control over the alignments, length distribution, token frequency distribution, and level of noise in the data.

All three synthetic tasks are based on a 1-to-1 paired dictionary of source and target synthetic tokens: $\mathcal{S}$ for source and $\mathcal{T}$ for target. We define a pairwise mapping between the two vocabularies such that each synthetic source token $\mathcal{S}_{i}$ is paired with a corresponding synthetic target token $\mathcal{T}_{i}$ for each $i \in 1 \ldots N$, where $N$ is the size of the paired vocabulary. In the examples below, the source vocabulary consists of all $26^3$ = 17576 three-character synthetic tokens that can be created using the lowercase English letters $\{ a, \ldots, z\}$.

\subsubsection{Synthetic Task 1: Identity Operation}
\label{task1:identity}

The simplest of the pre-training tasks we consider is the identity operation, which has been previously proposed by \citet{wu2022insights} as a synthetic task for language model pre-training. For this task, the source and target sentences are identical. We include it not because we believe it to be in any way a proxy for the true translation task, but instead to serve as the simplest possible baseline sequence-to-sequence synthetic task. We generate parallel sentence pairs by first sampling a sentence length $L$ from the normal distribution. Each source token $s_i$ for $i = 1 \ldots L$ is sampled uniformly from the source vocabulary $\mathcal{S}$. The target sentence is simply a copy of the source:

\begin{table}[!h]
\centering
\begin{footnotesize}
\begin{tabular}{r|l}
src & {\tt cea qne jda rnu jkq ozf dke kzl hpo} \\ 
trg & {\tt cea qne jda rnu jkq ozf dke kzl hpo} \\ 
\end{tabular}
\end{footnotesize}
\end{table}

\subsubsection{Synthetic Task 2: Case-Mapping}
\label{task2:casemap}

Our second pre-training task defines a case-mapping operation. Each synthetic parallel sentence pair consists of the same sequence of tokens but the source sentence is lowercase and the target sentence is uppercase. We also design an extension of this task that includes insertions and deletions. Source and target tokens can be deleted with fixed probability $d_s$ (for source) and $d_t$ (for target). Random insertions and deletions are added to avoid having identical source and target lengths for every sentence pair, which might entrench the tendency of the model to mimic such behavior even at the fine-tuning stage where it is likely inappropriate. From the perspective of the translation task, a sentence pair with a missing target token corresponds to a deletion, while a missing source token corresponds to an insertion. The following example shows a parallel sentence pair for the case-mapping task with fixed source and target deletion probabilities $d_s = d_t = 0.15$:

\begin{table}[!h]
\centering
\begin{footnotesize}
\begin{tabular}{l|l}
src & {\tt qdo zwj iub uxj pls nsn igk mrz ojw} \\ 
trg & {\tt QDO ZWJ IUB KWP UXJ PLS NSN IGK MRZ OJW} \\  
\end{tabular}
\end{footnotesize}
\end{table}

\subsubsection{Synthetic Task 3: Permuted Trees}
\label{task3:pbtrees}

The third of our synthetic pre-training tasks is designed to reflect some aspects of the reordering process that occurs during natural language translation. We first generate random sentences with normally distributed lengths and uniformly distributed synthetic tokens, as for tasks 1 and 2. We then induce an artificial binary tree over the source sentence by picking a random point at which to split the sentence, and recursively repeat this process for the left and right sub-strings. The resulting binary tree structure allows us to generate synthetic parallel data with reordering that preserves the alignment of contiguous source-to-target token spans. The target tree is generated as a permutation of the source tree: we randomly swap left and right sub-trees with some fixed probability $r$. Generating synthetic sentence pairs in this way implies the existence of lexicalised synchronous context free grammar (SCFG) rules \citep{chiang2007hierarchical} that could be used to generate the sentence pair as a parallel derivation. The example below shows a synthetic sentence pair generated using this method:

\begin{table}[!h]
\centering
\begin{footnotesize}
\begin{tabular}{r|l}
src & {\tt [ jtx [ [ urs [ ktp [ hme nmc ] ] ] pep ] ]} \\ 
trg & {\tt [ JTX [ [ URS [ [ HME NMC ] KTP ] ] PEP ] ]} \\  
\end{tabular}
\end{footnotesize}
\end{table}

\begin{figure}[!b]
\begin{center}
\includegraphics[width=0.8\linewidth]{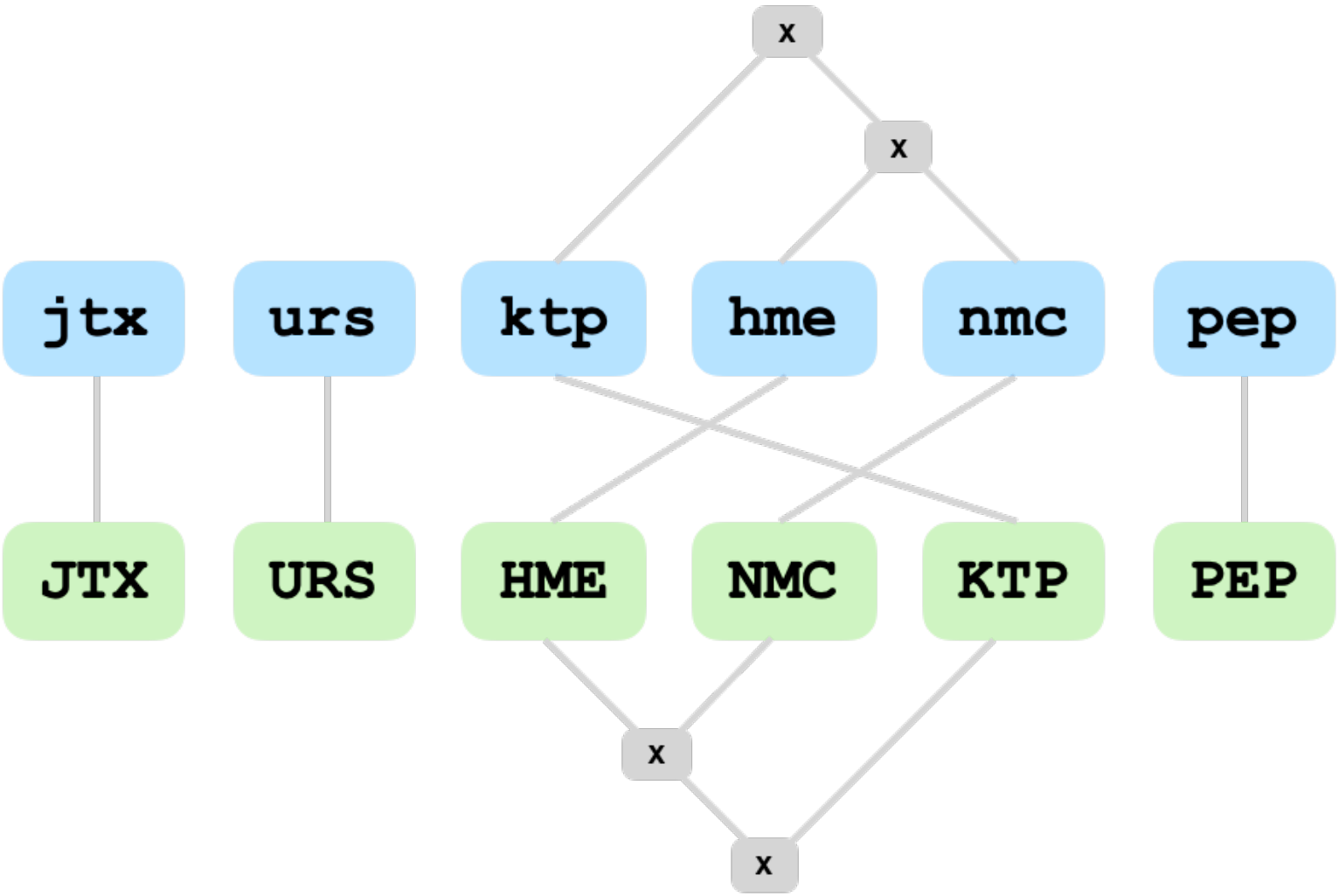}
\end{center}
\caption{Example synthetic sentence pair and partial derivation for the aligned permuted binary trees task. In this example, a single non-terminal node was reordered.
}
\label{fig:pb-trees-example}
\vspace{-1.0em}
\end{figure}

Parentheses indicating the tree structure are shown for clarity. During pre-training, however, only the source and target synthetic token sequences are actually seen by the model. In this example, the source token `{\tt ktp}' was reordered with respect to the sub-tree containing the tokens `{\tt hme nmc}'. Figure \ref{fig:pb-trees-example} shows the token-level alignment and reordering operations encoded by this parallel sentence pair.

\section{Experimental Framework}
\label{sec: experiment}
We evalute our synthetic pre-training data generation methods for NMT using using both English-centric and non-English-centric language pairs.

\subsection{Experiment Setup}
\label{sec:dataset-prepartion}

\paragraph{English-Centric Language Pairs}
For English-centric translation directions, we use fine-tuning data sets similar to \citet{aji2020neural}. For German-English, we use the official data from the WMT 2014 News Translation Task. For Myanmar-English, the fine-tuning data consists of 18.0k parallel sentence pairs in the news domain collected for the Asian Language Treebank (ALT) project \citep{ding2018nova}. We use the original train, dev and test split. For Indonesian-English, we use a filtered set of 24.6k parallel sentence pairs from the IDENTIC v1.0 corpus \citep{larasati2012identic} which covers various genres. We randomly divide the original corpus into distinct train (90\%), dev (5\%), and test (5\%) sets. For Turkish-English, we use data from the WMT 2017 News Translation Task \citep{yepes2017findings}. The training set includes 207.7k parallel sentence pairs. We use the WMT {\tt newsdev2016} set for validation, and report results on {\tt newstest2017}.

\paragraph{Non-English-Centric Language Pairs}
For non-English-centric directions, we simulate low-resource translation conditions by sampling data from OPUS NLP \citep{tiedemann2012parallel}. The non-English-centric language pairs we evaluate are as follows: Indonesian-Myanmar, Indonesian-Turkish, Indonesian-Tagalog, Myanmar-Turkish, Myanmar-Tagalog, Tagalog-Turkish, German-Indonesian, and German-Myanmar. For each pair, we simulate low-resource conditions by creating fine-tuning sets of size 10k, 25k, 50k, and 100k via sampling from the set of all parallel corpora for that language pair on OPUS NLP. Minimal filtering is applied to our parallel data sets: we remove duplicates, discard sentences with extreme length ratios, and keep only sentence pairs for which the {\tt fasttext} \citep{joulin2016fasttext} language ID matches the stated source and target.


\paragraph{Evaluation}
Following the evaluation setting of large-scale multilingual models such as FLORES-101 \cite{goyal2022flores}, we score our translation hypotheses using {\tt sentencepiece} BLEU \citep{papineni-etal-2002-bleu} (spBLEU). This avoids the need for custom post-processing for individual languages with unusual scripts and/or complex morphology such as Burmese.

\paragraph{Model Training Strategy}
Our experiments consist of a pre-training stage followed by a fine-tuning stage. We use the transformer sequence-to-sequence `base'  model architecture \citep{DBLP:journals/corr/VaswaniSPUJGKP17} for all translation experiments. Since our goal is to gain insight into the relative importance of various aspects of synthetic pre-training, our baseline models are created by fine-tuning randomly initialized models using only the downstream task parallel data.

We use {\tt fairseq} \citep{ott-fairseq-paper} to train our models with the Adam \cite{kingma2014adam} optimizer. We reset the learning rate scheduler and optimizer before starting the fine-tuning stage. Pre-training and fine-tuning continue until the BLEU score on the validation set converges. Further implementation details can be found in \cref{appendix:implementation_details}.

\subsection{Pre-training with Obfuscated Data}
\label{sec:obfuscated-pretraining-experiments}
Following previous work that showed German-to-English to be a good pre-training direction for several language pairs \citep{aji2020neural}, we also use German-to-English (\texttt{de-en}) for pre-training and randomly sample two million pairs from its training corpus to use as obfuscated parallel data. We vary the obfuscation ratio $R$ from 0\% to 100\% in 25\% increments. After pre-training, we fine-tune the models on the real-world parallel training corpus (described in \cref{sec:dataset-prepartion}) for each downstream language pair. We also investigate the scaling effect of different fine-tuning set sizes and show the results in Appendix \ref{appendix:obfuscation-scaling}.

\begin{figure}[t]
     \centering
    \includegraphics[width=\linewidth]{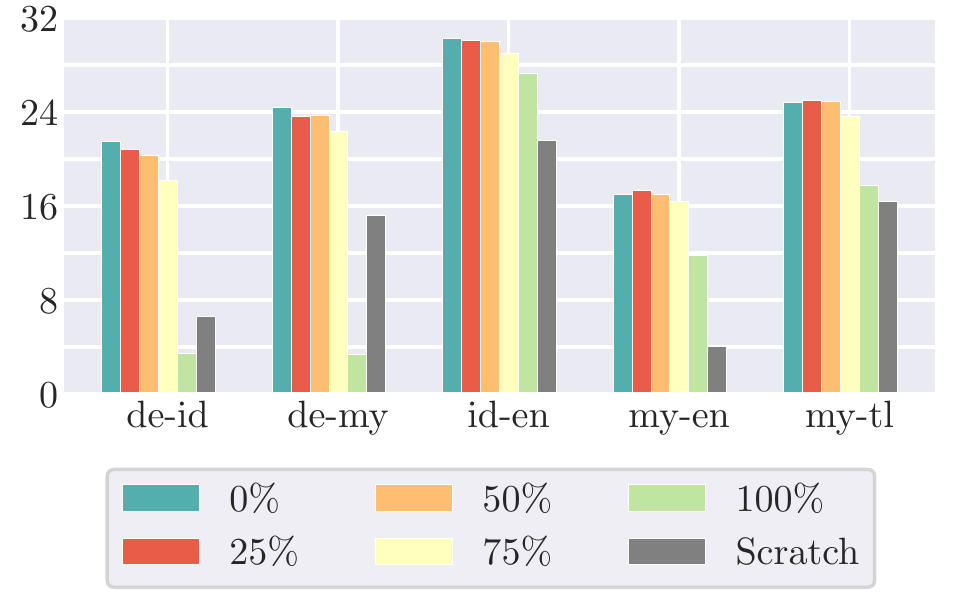}
    \label{fig:obfuscation unmatched on WMT}
     \vspace{-1.0em}
    \caption{Translation spBLEU scores after pre-training with different levels of obfuscation and real-world fine-tuning on downstream language pairs. \textit{Scratch} refers to training from scratch using only fine-tuning data. Similar results on FLORES can be found in ~\cref{appendix:obfuscation-more-results}.
     }
     \vspace{-1.0em}
\label{fig:obfuscation results}
\end{figure}

We report spBLEU scores on the test set for each language pair in \cref{fig:obfuscation results}. We find that, surprisingly, even when as much as 75\% of the pre-training data is obfuscated, the models are still able to achieve high or even comparable spBLEU scores to real-world pre-trained models (i.e., those with 0\% obfuscation). Additionally, most of the models pre-trained on obfuscated data performed better than those trained from scratch on real-world fine-tuning data, even when the pre-training data was 100\% obfuscated (e.g., 100\% in \texttt{id-en}, \texttt{my-en}, and \texttt{my-tl}). This suggests that a small proportion of real-world data can provide the majority of the benefits of large-scale regular pre-training, implying a promising research direction for efficient pre-training or improving low-resource NMT.

\subsection{Pre-training with Phrase Concatenation}
\label{sec:phrase-concatenation-experiments}

The translation decoding results in Table~\ref{table:synthetic-pretraining-results-english-centric} show substantial transfer learning benefits from pre-training with 2m sentence pairs of synthetic data generated by concatenating uniformly sampled aligned phrase pairs (phrase-cat). Compared to a model with no pre-training, i.e. one that trains from random initialization using only the fine-tuning data (random-init), we observe large gains of up to +9.9 spBLEU for language pairs with less than 25k of fine-tuning data ({\tt my}$\leftrightarrow${\tt en} and {\tt id}$\leftrightarrow${\tt en}). The gains of +1.4 to +2.1 for {\tt tr}$\leftrightarrow${\tt en} are smaller: this pair has more fine-tuning data (>200k pairs) so the improved coverage and robustness of synthetic pre-training is less critical for good performance. It is important to note that this method does not utilize any additional real parallel or monolingual data, but instead derives new data directly from the existing fine-tuning corpus. Our synthetic pre-training corpus, although unnatural at the sentence-level, contains many phrase-level alignments and reordering information which reinforces the translation knowledge captured by the model. Any destructive effect from presenting to the model during pre-training sentence pairs with unnatural word order or bad grammar, can be rectified in the fine-tuning stage by showing the model the original fluent source and target sentences.

\subsection{Pre-Training with Synthetic Data}
\label{sec:synthetic-tasks-experiments}

We pre-train transformer \citep{DBLP:journals/corr/VaswaniSPUJGKP17} models using 2m sentence pairs of synthetic parallel data to match the data size used in our obfuscation experiments. We further explore the effect of scaling the synthetic pre-training data size in \cref{appendix:synthetic-Data-Scaling}. Separate synthetic training sets were generated for each of the three task variants described in Section \ref{section:synthetic-pretraining-tasks}. Additional sets of 4000 synthetic pairs were generated as validation data. Each pre-trained model is subsequently fine-tuned with real parallel data for a specific language pair: {\tt my}$\leftrightarrow${\tt en},  {\tt id}$\leftrightarrow${\tt en}, and {\tt tr}$\leftrightarrow${\tt en}. In Table \ref{table:synthetic-pretraining-results-english-centric}, we report {\tt sentencepiece} BLEU (spBLEU) \cite{goyal2022flores} scores for our three synthetic pre-training task variants. For comparison purposes, we also show the scores obtained without pre-training -- i.e.\ a randomly initialized model trained on only the fine-tuning data.
\begin{table*}[t]
\centering
\resizebox{\textwidth}{!}{%
{\small
\begin{tabular}{crrrrrrrrrrrr}
\toprule
\multirow{2}{*}{} &
  \multicolumn{2}{c}{{\tt my-en}} &
  \multicolumn{2}{c}{{\tt id-en}} &
  \multicolumn{2}{c}{{\tt tr-en}} &
  \multicolumn{2}{c}{{\tt en-my}} &
  \multicolumn{2}{c}{{\tt en-id}} &
  \multicolumn{2}{c}{{\tt en-tr}} \\ \cmidrule(lr){2-13} 
                                & Test & Flores & Test & Flores & Test & Flores & Test & Flores & Test & Flores & Test & Flores \\ \midrule
\multicolumn{1}{l}{scratch}    & 4.1  & 1.8    & 18.2 & 7.2    & 14.7 & 17.7   & 16.2 & 6.3    & 19.1 & 8.3    & 17.0 & 16.4   \\ \midrule
\multicolumn{1}{l}{identity}   & 3.2  & 1.1    & 16.8 & 7.6    & 12.4 & 13.8   & 12.7 & 4.5    & 18.1 & 9.7    & 13.8 & 13.5   \\
\multicolumn{1}{l}{case-map}   & 6.7  & 1.6    & 21.8 & 12.1   & 13.4 & 15.1   & 16.4 & 6.0    & 22.9 & 13.8   & 15.6 & 15.2   \\
\multicolumn{1}{l}{pb-trees}   & 11.4 & 2.5    & 23.1 & 12.2   & 14.4 & 16.9   & 18.9 & 7.0    & 23.8 & 14.4   & 16.6 & 16.3   \\ \midrule
\multicolumn{1}{l}{phrase-cat} & 14.0 & 3.9   & 27.3 & 14.4   & 16.5 & 19.1   & 23.0 & 8.6    & 28.1 & 17.0   & 18.4 & 18.5   \\ \bottomrule
\end{tabular}%
}
}
\caption{Translation decoding results (spBLEU) for three purely synthetic pre-training variants and concatenation of aligned phrases vs. fine-tuning from a randomly initialized baseline (scratch) (English-centric language pairs).}
\vspace{-1em}
\label{table:synthetic-pretraining-results-english-centric}
\end{table*}
Our first observation is that synthetic pre-training with the identity operation task (\S\ref{task1:identity}) does not perform well. For all three language pairs it is slightly worse than simply fine-tuning from a randomly initialized model. This is to be expected since the pre-training task is too crude: a simple copy operation from source to target with identical lengths. Pre-training with the case-mapping synthetic task (\S\ref{task2:casemap}) and deletion probability $d_s=d_t=0$ improves the scores, with gains of +1.0 to +5.0 spBLEU over the identity operation on our test set. Although the case-mapping pre-training task is still quite crude, it is able to beat fine-tuning from a randomly initialized model for both Myanmar-to-English and Indonesian-to-English. Our best performing synthetic task is pb-trees (\S\ref{task3:pbtrees}) with a node reordering probability $r=0.15$. This model shows that transfer learning from synthetic pre-training to real-world tasks can be substantial, with scores as high as +7.3 spBLEU over the baseline for Myanmar-to-English and +4.9 for Indonesian-to-English. We do not see gains for Turkish-to-English for any of our purely synthetic pre-training tasks. The fine-tuning data for this language pair is much larger than that of the other language pairs. As the fine-tuning data size increases, the benefits of transfer learning from pre-training diminish.

We also evaluate the strongest of our three purely synthetic pre-training tasks, pb-trees, on additional non-English-centric language pairs. \cref{table:synthetic-pretraining-additional-pairs} in \cref{appendix:supplementary-pretraining-results} shows spBLEU decoding results for these additional pairs. 
We compare performance over a range of different fine-tuning set sizes. On both OPUS-Test and FLORES-devtest, and for the majority of fine-tuning set sizes, synthetic pre-training with the pb-trees task typically outperforms fine-tuning from a randomly initialized baseline.

\section{Analysis and Discussion}
\label{sec: analysis_and_discussion}

\subsection{Synthetic Knowledge Transfer}
\label{discussion:what-is-actually-transferred}

In this section, we discuss what kind of useful representations are actually learned by the model when pre-training with purely synthetic tasks and data. Our empirical study has shown that pre-training on synthetic data can result in improved translation quality after fine-tuning for a specific language pair. Even though the pre-training data is entirely synthetic, the model must have successfully learned representations and structures relevant for translation that can be leveraged via transfer learning to the downstream task.

In Table \ref{table:wordpiece-overlap-statistics}, we show the word piece overlap between our tokenized synthetic pre-training corpus and the real human language corpus for three fine-tuning language pairs. Our vocabulary consists of $26^3$ paired lowercase-uppercase synthetic tokens, but after tokenization the number of unique word pieces is much lower. For example, there are only 3,541 unique source and 2,405 unique target word pieces after tokenizing a corpus of 2M synthetic parallel sentence pairs. The fine-tuning data, although much smaller, has a far greater token diversity for English, Indonesian, and Turkish. Myanmar is the exception: it is aggressively segmented by the XLMR {\tt sentencepiece} model which results in far fewer unique word pieces.

We compute the intersection between the set of word pieces in the synthetic pre-training data and those in the fine-tuning data in the last column of \cref{table:wordpiece-overlap-statistics}. We observe low word piece overlap. For example, only 35 of the 3541 word pieces that occur in the source side of the synthetic corpus also occur in the source side of the {\tt my-en} corpus. This number is low because the Myanmar script is so different from English. But overlap remains low even for languages such as Indonesian and Turkish which have similar alphabets to English. Low levels of overlap were also observed in our obfuscated pre-training experiments (\cref{tab: obfuscation_word_overlap_de-en}). The low word piece overlap means that most of the word embeddings learned during pre-training have little relevance to the fine-tuning or inference stages. We conclude that any transfer learning benefit exhibited by the model on the downstream task must be captured in the inner layers of the transformer.


\begin{table}[]
\centering
{\small
\begin{tabular}{ccrrr}
\toprule
Pair & PT/FT & $|V_{PT}|$ & $|V_{FT}|$ & Overlap \\ \midrule
\multirow{2}{*}{{\tt my-en}} & src: {\tt lc/my} & 3,541 &  1,598 &    35 \\
& trg: {\tt uc/en} & 2,405 & 18,514 &   740 \\ \midrule
\multirow{2}{*}{{\tt id-en}} & src: {\tt lc/id} & 3,541 & 18,095 & 1,377 \\
& trg: {\tt uc/en} & 2,405 & 18,167 &   740 \\ \midrule
\multirow{2}{*}{{\tt tr-en}} & src: {\tt lc/tr} & 3,541 & 24,616 & 1,938 \\
& trg: {\tt uc/en} & 2,405 & 26,236 & 1,358 \\ \bottomrule
\end{tabular}
}
\caption{Tokenized pre-training (PT) and fine-tuning (FT) word piece counts and overlap statistics: `{\tt lc}' and `{\tt uc}' denote lowercase and uppercase synthetic tokens.}
\vspace{-1em}
\label{table:wordpiece-overlap-statistics}
\end{table}


\subsection{Lexical and Structural Knowledge}
\label{sec:lexical-and-structural-comparison}

The results in \cref{table:synthetic-pretraining-results-english-centric} show phrase-cat to be an effective pre-training strategy for low-resource NMT. Both lexical and structural knowledge is captured in the aligned phrases. However, since the phrases are sampled from the uniform distribution, long-distance structure is ignored and only local reordering information is captured. The pb-trees method also allows us to encode structural knowledge into our synthetic data since it is possible to generate sentence pairs that reorder sub-trees over long distances. Comparing the effectiveness of both methods shows that surprising gains in translation quality are possible even for synthetic data generation methods such as phrase-cat that encode only very local structural knowledge. This insight, that it is mainly collocations (especially, for NMT, parallel collocations) agrees with the conclusions about the relative lack of importance of word order to LM pre-training in \citet{sinha2021masked}.

\subsection{Translation Quality vs. Toxicity}
\begin{table*}[t]
\centering
\resizebox{\textwidth}{!}{%
\begin{tabular}{@{}lcrrrrrrrrrr@{}}
\toprule
\multicolumn{1}{c}{\multirow{2}{*}{}} &
  \multirow{2}{*}{{Model}} &
  \multicolumn{2}{c}{\tt{de-id}} &
  \multicolumn{2}{c}{\tt{de-my}} &
  \multicolumn{2}{c}{\tt{id-en}} &
  \multicolumn{2}{c}{\tt{my-en}} &
  \multicolumn{2}{c}{\tt{my-tl}} \\ \cmidrule(l){3-12} 
\multicolumn{1}{c}{}          &                                  & BLEU & Toxicity & BLEU & Toxicity & BLEU & Toxicity & BLEU & Toxicity & BLEU & Toxicity \\ \midrule
\multicolumn{1}{c|}{{Baseline}} & \multicolumn{1}{c|}{{scratch}}     & 6.6  & 0.68     & 15.2 & 0.01     & 18.2 & 0.05     & 4.1  & 0.02     & 16.4 & 0.04     \\ \midrule
\multicolumn{1}{c|}{\multirow{2}{*}{\begin{tabular}[c]{@{}c@{}}{Large Pretrained}\\ {Multilingual Model}\end{tabular}}} &
  \multicolumn{1}{c|}{{M2M-100}} &
  32.9 &
  0.68 &
  9.1 &
  0.03 &
  30.2 &
  0.28 &
  1.8 &
  0.15 &
  14.2 &
  0.06 \\
\multicolumn{1}{l|}{}         & \multicolumn{1}{c|}{{FLORES-101}} & 30.0   & 0.63     & 12.3 & 0.03     & 26.0   & 0.23     & 4.6  & 0.18     & 12.8 & 0.08     \\ \midrule
\multicolumn{1}{c|}{\multirow{3}{*}{\begin{tabular}[c]{@{}c@{}}{Synthetic}\\ {Pre-training}\end{tabular}}} &
  \multicolumn{1}{l|}{{obfuscation}} &
  18.2 &
  0.34 &
  22.4 &
  0.01 &
  29.0 &
  0.11 &
  16.4 &
  0.08 &
  23.6 &
  0.04 \\
\multicolumn{1}{c|}{}         & \multicolumn{1}{c|}{{phrase-cat}}  & 14.7 & 0.50     & 19.6 & 0.02     & 27.3 & 0.10     & 14.0 & 0.02     & 22.5 & 0.03     \\
\multicolumn{1}{c|}{}         & \multicolumn{1}{c|}{{pb-trees }   }                      & 11.7 & 0.45     & 12.3 & 0.01     & 23.1 & 0.10     & 11.4 & 0.01     & 20.7 & 0.02     \\ \bottomrule
\end{tabular}%
}
\caption{BLEU scores and toxicity rates for various models on low-resource language pairs. Baseline is training on fine-tune real-world data as lower bound of performance. Large pre-trained models are upper bound of performance.}\vspace{-1.0em}
\label{tab: All_models_toxicity_bleu}
\end{table*}



To evaluate model toxicity, we consider catastrophic mistranslations \citep{costa2022no}. These errors occur when a model hallucinates toxic terms in the translated text, even though no such terms occur in the source text. Following the toxicity measurement setup of \citet{goyal2022flores}, we use the FLORES Toxicity-200\footnote{\url{http://github.com/facebookresearch/flores/tree/main/toxicity}} word lists to calculate the toxicity rate of translations produced by a model. The lists cover 200 languages and contain frequently used profanities, insults, and hate speech terms. We consider a sentence toxic if it contains words that match entries in these lists. The toxicity rate for each model is defined as the proportion of sentences with hallucinated toxicity in translations of the test set and a larger set of 100k monolingual sentences randomly sampled from CC-100 \citep{wenzek-etal-2020-ccnet,conneau-xlmr-paper}. We compare BLEU scores and toxicity rates for various models including current state-of-the-art large pre-trained multilingual translation models in \cref{tab: All_models_toxicity_bleu}.

\paragraph{Results and Analysis} 
We first observe that models pre-trained on synthetic data obtain significantly higher BLEU scores than baselines trained from scratch using only the fine-tuning data. This confirms that our proposed synthetic tasks indeed capture useful knowledge that can be applied through transfer learning to low-resource NMT tasks. 
When compared to the multilingual translation models FLORES-101 (615M parameters) and M2M-100 (1.2B parameters), we note that models pre-trained on synthetic data obtain comparable performance for languages {\tt my-en} and even outperform multilingual models by a large margin on {\tt de-my}, {\tt id-en}, and {\tt my-tl}, though with inferior translation quality on \texttt{de-id}. 
It should be noted that some of these language pairs represent zero-shot directions for M2M-100. We compare our synthetic methods with the standard NMT data augmentation technique of back-translation in \cref{appendix:back-trans}.

While these results are quite promising, we note that our goal in this paper is not to surpass the state-of-the-art in translation quality achieved by large-scale massively multilingual models on low-resource NMT. Instead, we seek to further understand which properties of pre-training based on synthetic tasks and data - along the structural and lexical knowledge axes of \cref{fig:three-approaches} - enhance transfer learning performance, while minimizing toxicity and other data issues inherent in models that rely on large-scale pre-training using real data.  

Analyzing toxicity, we observe the presence of catastrophic mistranslations in all models, but less frequently when training from scratch in most cases. This is because the low-resource fine-tuning data contains very little toxic content. On the other hand, as noted above, the BLEU scores when training models from scratch are very low. We see that the FLORES-101 and M2M-100 models both exhibit toxicity, since they were pre-trained on real-world corpora that can include toxic content. Our results show that synthetic pre-training can produce models with comparable BLEU scores while significantly reducing catastrophic mistranslations. 
We observe that parallel data generated from permuted binary trees has the lowest toxicity among the three synthetic pre-training methods, since it relies on purely synthetic data. This may indicate that patterns in the data can still trigger toxic terms, even after the words have been obfuscated or phrases have been shuffled. We include additional toxicity results and analysis in  \cref{sec:appendix_toxicity}.

\section{Conclusion}
\label{sec:conclusion}
Our study of synthetic pre-training tasks for NMT showed that pre-training benefits can still be achieved even when using synthetic or obfuscated data. Additionally, we have shown that synthetic data has the potential to reduce model toxicity compared to models trained on web-scale crawled corpora. Our research provides insights into what types of knowledge transfer make for a good pre-trained model. We believe that synthetic data augmentation techniques based on synthetic tasks and procedurally generated data are a promising solution for addressing pre-training data concerns, and can lead to efficient, accurate, and trustworthy NMT. In future work, we plan to further investigate synthetic pre-training by exploring more advanced data generation models and directly optimizing the parameters for specific downstream fine-tuning tasks. Increasing the effectiveness of synthetic data at different data scales is also worthy of further exploration.


 

\section{Limitations}
Our work seeks to gain insight into what pre-training knowledge is transferred and useful for downstream fine-tuning in NMT using synthetic tasks and data. We note that changes in the data generation methods do require re-running the pre-training stage, which is computationally expensive compared to the fine-tuning stage.


Our current synthetic data generation methods are somewhat crude. Although they are designed to encode varying degrees of lexical and structural translation knowledge, they do so in a rather simplistic way. For example, sampling phrases from the normal distribution ignores distributional frequencies which represent information that is likely useful for the synthetic data generation task. In this paper we have presented some interesting initial findings regarding the suitability of synthetic pre-training for NMT. We plan to explore more sophisticated data generation models in future work.

We acknowledge that synthetic pre-training is unlikely to surpass the quality of real-world massively multilingual pre-trained models in performance, especially if synthetic data is the only data used for pre-training. However, good performance can probably be achieved by combining synthetic pre-training and real-data pre-training. Of course, this risks exposing the model to toxic and sensitive or private content. Therefore, concerns of both model quality and data quality should be considered when evaluating the impact and benefits of synthetic pre-training. We view synthetic pre-training as a complimentary approach to finding an optimal balance rather than as a replacement for previous state-of-the-art NMT pre-training methods.

\section{Ethics Statement}
All of the training data used in our experiments are official releases of publicly available benchmarks. In addition, the toxic word lists used to measure toxicity are obtained from the public FLORES repository which requires a password to access, thus reducing the risk of hacking by a malicious user or adversarial bot. In addition, as for the issue of hallucinated toxicity discussed previously, we note that our work also has the potential to address other problematic translation behaviors, such as hallucinated bias. 

\section{Acknowledgements}
This material is based upon work supported by the Defense Advanced Research Projects Agency under Contract No. FA8750-19-C-1001. Disclaimer: Any opinions, findings and conclusions or recommendations expressed in this material are those of the author(s) and do not necessarily reflect the views of the Defense Advanced Research Projects Agency. Zexue He is supported by an IBM Ph.D. Fellowship and is independent of the Defense Advanced Research Projects Agency.




\bibliographystyle{acl_natbib}
\bibliography{anthology,custom}

\newpage

\appendix

\clearpage
\section{Supplementary Results}

\subsection{Scaling Effect of Obfuscated Pre-training}
\label{appendix:obfuscation-scaling}

\begin{figure}[b]
     \centering
         \includegraphics[width=\linewidth]{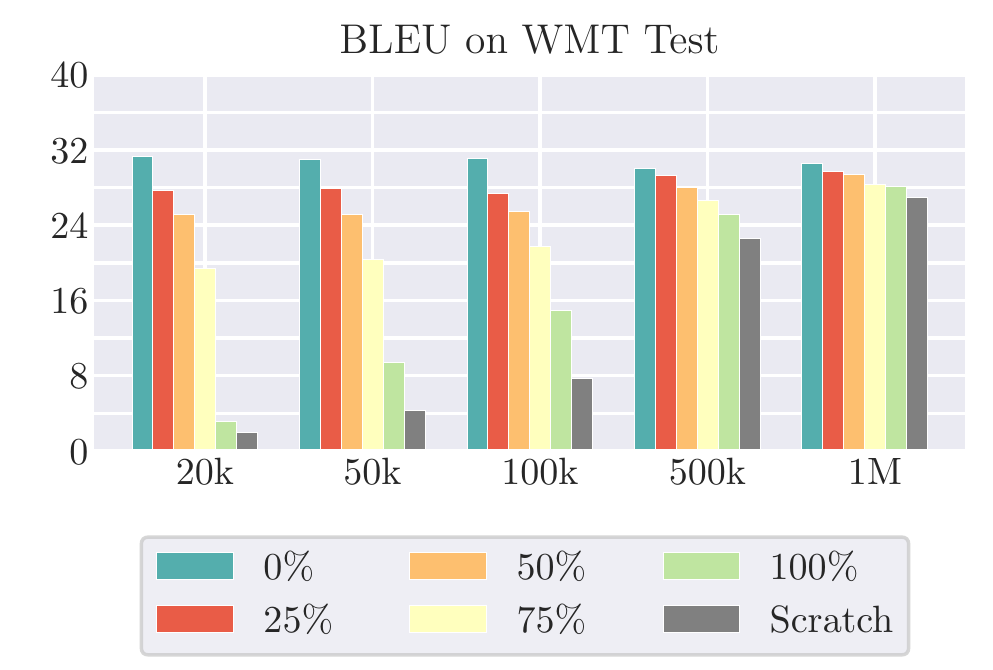}

    \caption{Translation results after pre-training with different levels of obfuscation and real-world fine-tuning on the same language pairs, with various quantities of  fine-tuning data in \texttt{de-en}. \textit{Scratch} refers to training from scratch using only fine-tuning data.
     }
     \vspace{-1.0em}
      \label{fig:obfuscation de-en on WMT} 
\end{figure}

We first evaluate the performance of regular pre-training and fine-tuning with various quantities of real-world German-to-English data. The results in \cref{fig:obfuscation de-en on WMT} show that the highest BLEU scores are obtained by using regular real-world parallel data (i.e.\ 0\% obfuscation). We compare vs. models trained solely on the fine-tuning data (`Scratch'): the resulting BLEU scores are quite low when the training data size is small, confirming the importance and benefits of pre-training for NMT.


\subsection{FLORES Obfuscated Pre-training Results}
\label{appendix:obfuscation-more-results}
\begin{figure*}[t]
     \centering
     \begin{subfigure}[b]{0.45\textwidth}
         \centering
         \includegraphics[width=\textwidth]{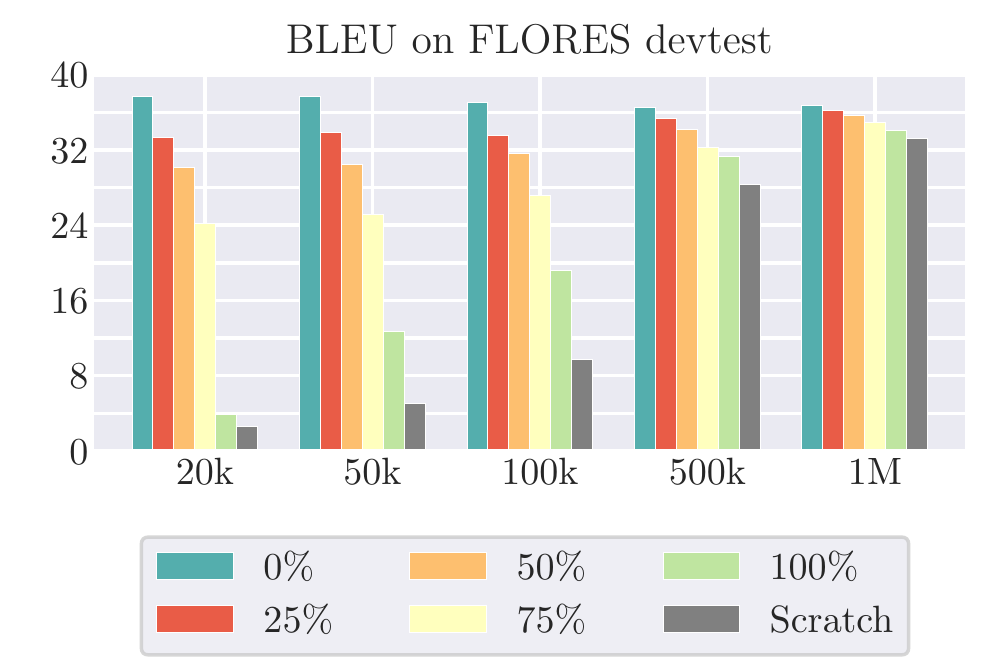}
         \caption{}
         \label{fig:obfuscation de-en on flores}
     \end{subfigure}
     \hfill
     \begin{subfigure}[b]{0.45\textwidth}
         \centering
         \includegraphics[width=\textwidth]{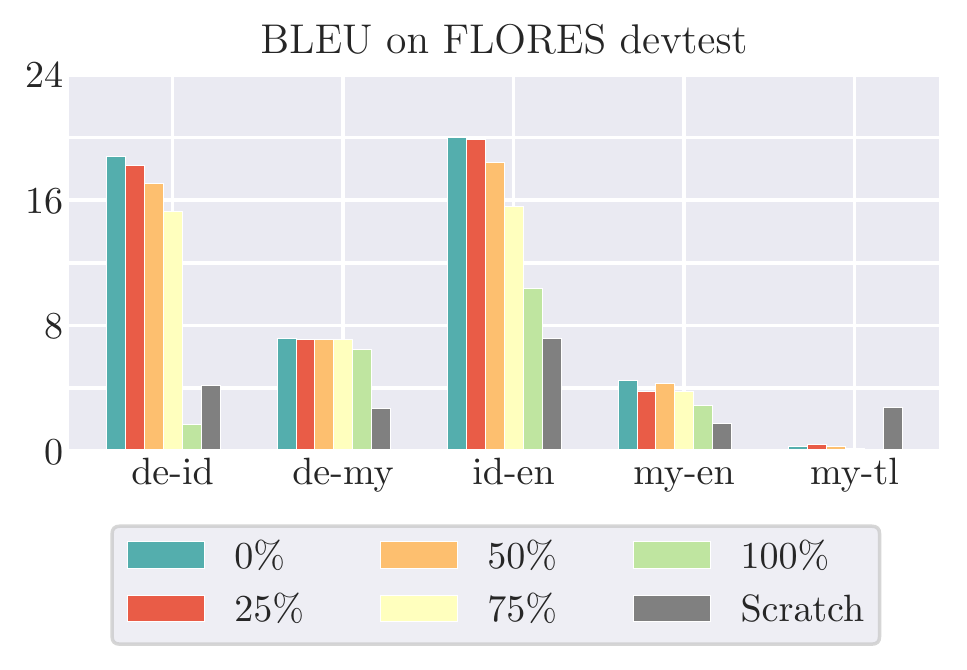}
         \caption{}
         \label{fig:obfuscation unmatched on flores}
     \end{subfigure}
     \caption{Translation decoding results on WMT for (a) regular parallel corpus (0\%) vs.\ obfuscated pre-training as a function of fine-tuning set size (x-axis) and obfuscation ratio (in different colors), and (b) unmatch conditions. 
     }
        \label{fig:three graphs}
\end{figure*}

We show additional decoding results for the matched (with source and target fine-tuning languages that are the same as the pre-training languages: \texttt{de-en}) vs. unmatched (with source or target fine-tuning languages that differ from the pre-training languages: \texttt{de-id, de-my, id-en, my-en, my-tl}) conditions of obfuscated pre-training on the FLORES devtest set in \cref{fig:three graphs}.

\subsection{Back-Translation Comparison}
\label{appendix:back-trans}
Back-translation \cite{sennrich2016improving} is an effective technique for improving the quality of machine translation. It works by creating new parallel sentence pairs by translating target-side monolingual data into the source language using an inverse direction MT system. The new sentence pairs consist of a (possibly noisy) back-translated source sentence paired with a high-quality target sentence. We compare our synthetic training methods to an NMT system that has been trained on an augmented data set that includes back-translated parallel data. We use our baseline models for \texttt{my-en} and \texttt{en-my} to produce the back-translated sentences. For each direction {\tt my-en} and {\tt en-my}, we generate an additional set of 2m back-translated sentences. The results are shown in \cref{tab:back-trans}. We note that back-translation provides only limited improvements vs. the baseline model trained from scratch for \texttt{my-en} and actually hurts for {\tt en-my}. This is because back-translation requires a good quality model in the target-to-source direction in order to produce accurate and relevant translations. The {\tt my-en} baseline model is not of sufficiently high quality to produce useful back-translations. Our synthetic methods significantly outperform back-translation for both translation directions, confirming our expectation about the limitations of back-translation in low-resource conditions, and further illustrating the potential of our proposed synthetic approaches.

\begin{table}[]
\centering
\resizebox{0.85\linewidth}{!}{%
\begin{tabular}{@{}lrrrr@{}}
\toprule             & \multicolumn{2}{c}{\tt{my-en}}     & \multicolumn{2}{c}{\tt{en-my}}    \\ \cmidrule(l){2-5}
Model            & Test          & Flores        & Test         & Flores       \\ \midrule
scratch          & 4.1           & 1.8           & 16.2          & 6.3          \\
back-translation & 10.7          & 2.0           & 11.1          & 4.1          \\
phase-cat        & 14.0 & 3.9 & 23.0 & 8.6 \\
pb-trees          & 11.4          & 2.5           & 18.9          & 7.0          \\ \bottomrule
\end{tabular}%
}
\caption{Synthetic pre-training v.s. back-translation on WMT test set and FLORES devtest set.}
\label{tab:back-trans}
\end{table}

\subsection{Synthetic Pre-training Data Scaling}
\label{appendix:synthetic-Data-Scaling}
\begin{figure}[t]
\begin{center}
\includegraphics[width=1.0\linewidth]{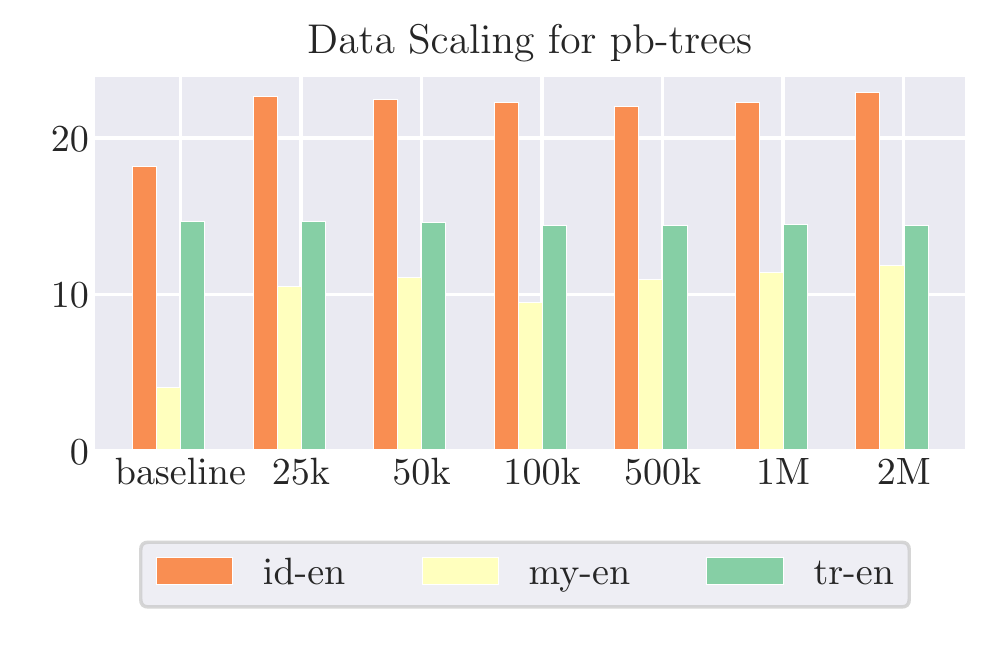}
\par\bigskip
\includegraphics[width=1.0\linewidth]{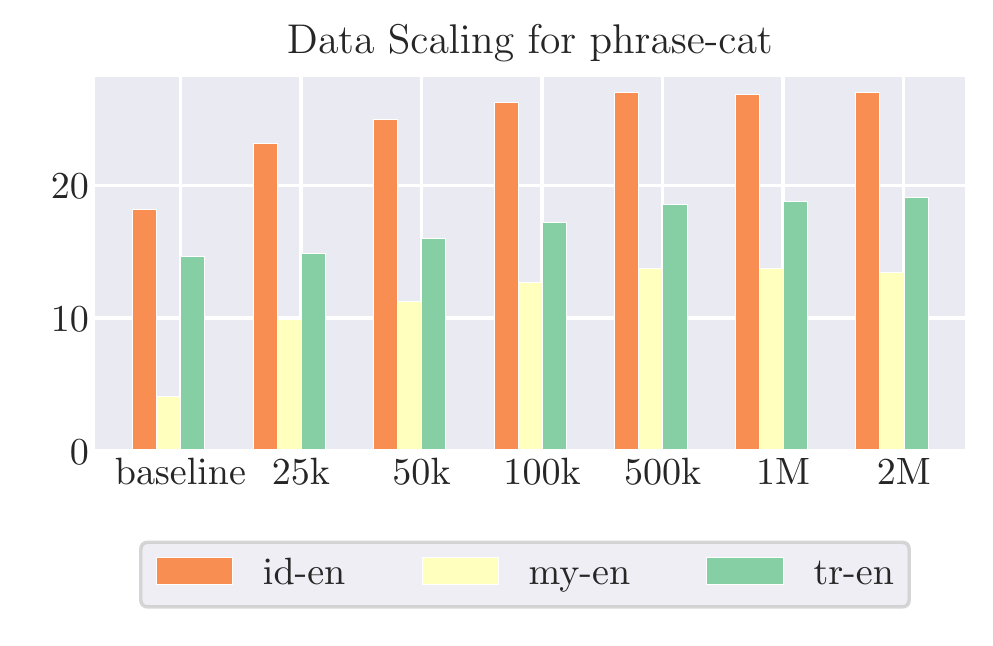}
\end{center}
\caption{Effect on BLEU score of scaling up the size of the procedurally generated parallel data used during pre-training for two of our synthetic tasks: permuted binary trees `pb-trees` (top), and concatenated aligned phrases `phrase-cat` (bottom).}
\label{fig:data-scaling-plots}
\vspace{-1.0em}
\end{figure}

\cref{fig:data-scaling-plots} shows the data scaling behavior of the pb-trees and phrase-cat synthetic pre-training methods. We pre-train each model with proper subsets of varying sizes sampled from the full 2m pairs used in the rest of our experiments. For pb-trees, the scaling is mostly flat. The BLEU scores, while consistently higher than the baseline (which uses no pre-training at all), increase only gradually with additional synthetic training data. The BLEU gains over the baseline are therefore a result of priming the model for the task of translation, rather than learning any useful real-world lexical relationships between the source and target languages. For phrase-cat, the data scaling curve is much more pronounced. For all three tasks, we observe steadily increasing BLEU scores with larger synthetic training set sizes, reaching a plateau at around 1m pairs. The phrase-cat method benefits from additional samples and combinations of real phrase pairs since the synthetic pairs provide additional coverage of possible word orders and translation relationships that can aid the subsequent fine-tuning and decoding of the testset.

\subsection{Further Analysis of Toxicity}
\label{sec:appendix_toxicity}
We further analyze the toxicity of our models by comparing the toxicity rate of source language sentences and their translations. Firstly, we test {\tt de-en} translation systems with obfuscated pre-training on WMT test, as shown in  \cref{tab: obfuscated_toxicity_de-id}. We observe that training with real-world data (i.e.\ obfuscation ratio $R$ = 0\%)  generates translations that contain toxic terms more frequently than they occur in the source. This indicates a toxicity amplification effect, a problem highlighted previously for toxicity \citep{costa2022no} and bias \citep{leino2018feature}. Pre-training with obfuscated data, however, is a promising way of mitigating this phenomenon, as shown by the big reduction in toxicity rate as the obfuscation ratio is increased. We observe a similar pattern for CC-100 data as well. The sentences in the CC-100 corpus are more toxic than those in the WMT testset (0.57\% $>$ 0.43\%). 

\begin{table*}[t]
\begin{minipage}[c]{0.5\textwidth}
\centering
\resizebox{0.9\linewidth}{!}{%
\begin{tabular}{cccccc}
\toprule
 &
  \multicolumn{5}{c}{\textbf{Obfuscation Ratio}} \\ \cmidrule(l){2-6} 
\multirow{-2}{*}{\begin{tabular}[c]{@{}c@{}}\textbf{Fine-Tuning}\\\textbf{Set Size}\end{tabular}} &
  \textbf{0\%} &
  \textbf{25\%} &
  \textbf{50\%} &
  \textbf{75\%} &
  \textbf{100\%} \\ \cmidrule(l){1-6} 
\textbf{20k} &
  \cellcolor[HTML]{E67C73}0.57 &
  \cellcolor[HTML]{FFFFFF}0.40 &
  \cellcolor[HTML]{FBE5E4}0.43 &
  \cellcolor[HTML]{F1F9F5}0.37 &
  \cellcolor[HTML]{57BB8A}0.00\\
\textbf{50k} &
  \cellcolor[HTML]{FBE5E4}0.43 &
  \cellcolor[HTML]{EB978F}0.53 &
  \cellcolor[HTML]{F6CBC8}0.47 &
  \cellcolor[HTML]{FFFFFF}0.40 &
  \cellcolor[HTML]{65C093}0.03 \\
\textbf{100k} &
  \cellcolor[HTML]{EB978F}0.53 &
  \cellcolor[HTML]{E3F3EB}0.33 &
  \cellcolor[HTML]{FFFFFF}0.40 &
  \cellcolor[HTML]{C7E8D8}0.27 &
  \cellcolor[HTML]{73C69D}0.07 \\
\textbf{500k} &
  \cellcolor[HTML]{F0B1AB}0.50 &
  \cellcolor[HTML]{F0B1AB}0.50 &
  \cellcolor[HTML]{E3F3EB}0.33 &
  \cellcolor[HTML]{E3F3EB}0.33 &
  \cellcolor[HTML]{FFFFFF}0.40 \\
\textbf{1M} &
  \cellcolor[HTML]{E67C73}0.57 &
  \cellcolor[HTML]{F6CBC8}0.47 &
  \cellcolor[HTML]{FFFFFF}0.40 &
  \cellcolor[HTML]{F1F9F5}0.37 &
  \cellcolor[HTML]{F1F9F5}0.37 \\ \bottomrule
\end{tabular}%
}
\end{minipage}
\begin{minipage}[c]{0.5\textwidth}
\centering
\resizebox{0.9\linewidth}{!}{%
\begin{tabular}{cccccc}
\toprule
 &
  \multicolumn{5}{c}{\textbf{Obfuscation Ratio}} \\ \cmidrule(l){2-6} 
\multirow{-2}{*}{\textbf{\begin{tabular}[c]{@{}c@{}}Fine-Tuning\\Set Size\end{tabular}}} &
  \textbf{0\%} &
  \textbf{25\%} &
  \textbf{50\%} &
  \textbf{75\%} &
  \textbf{100\%} \\ \midrule
\textbf{20k} &
  \cellcolor[HTML]{DFF2E9}0.37 &
  \cellcolor[HTML]{D0ECDE}0.33 &
  \cellcolor[HTML]{CFEBDE}0.33 &
  \cellcolor[HTML]{A3D9BF}0.21 &
  \cellcolor[HTML]{57BB8A}0.01 \\
\textbf{50k} &
  \cellcolor[HTML]{DFF2E9}0.37 &
  \cellcolor[HTML]{D7EFE3}0.35 &
  \cellcolor[HTML]{DEF1E8}0.37 &
  \cellcolor[HTML]{B7E2CD}0.26 &
  \cellcolor[HTML]{66C194}0.05 \\
\textbf{100k} &
  \cellcolor[HTML]{F5FAF8}0.43 &
  \cellcolor[HTML]{CDEBDC}0.32 &
  \cellcolor[HTML]{C4E7D5}0.30 &
  \cellcolor[HTML]{AADCC3}0.23 &
  \cellcolor[HTML]{95D4B5}0.17 \\
\textbf{500k} &
  \cellcolor[HTML]{DAF0E5}0.36 &
  \cellcolor[HTML]{E3F4EC}0.38 &
  \cellcolor[HTML]{D9F0E5}0.36 &
  \cellcolor[HTML]{CCEADC}0.32 &
  \cellcolor[HTML]{BBE3CF}0.27 \\
\textbf{1M} &
  \cellcolor[HTML]{E2F3EB}0.38 &
  \cellcolor[HTML]{FFFFFF}0.45 &
  \cellcolor[HTML]{DAF0E5}0.36 &
  \cellcolor[HTML]{D9EFE4}0.35 &
  \cellcolor[HTML]{CEEBDD}0.33 \\ \bottomrule
\end{tabular}%
}
\end{minipage}
\caption{Toxicity rate (\%) on WMT Test (left) and sampled CC-100 data (right). Results that increase toxicity compared to the source (0.43\% for WMT and 0.57\% for CC-100) are colored in red; otherwise they are colored in green. The degree of toxicity is shown by the darkness of the color.}
\label{tab: obfuscated_toxicity_de-id}
\end{table*}

\subsection{Word-Piece Overlap Statistics for Obfuscated Pre-Training}
Similar to \cref{discussion:what-is-actually-transferred}, we also report the token overlap between completely encrypted pre-training data (both source and target corpus) and real-world fine-tuning data, on {\tt de-en} as shown in \cref{tab: obfuscated_toxicity_de-id} and other language directions {\tt id-en}, {\tt my-tn}, and {\tt tr-en} in \cref{tab: obfuscation_word_overlap_other}. In {\tt de-en} translation, we notice that the overlap is just 0.08\% on the source language and 0.04\% on the target language, with the largest size of the fine-tuning set (1M). On low-resource language pairs, we can see there is almost no overlap between pre-training and fine-tuning on both source and target sides, as shown in \cref{tab: obfuscation_word_overlap_other}. This strong evidence supports the conclusion mentioned in  \cref{discussion:what-is-actually-transferred} -- most of the representations in the first layers are not touched during pre-training, and the benefits from pre-training may come from the inner layers which capture the transferable high-level knowledge for downstream tasks. 

\begin{table*}[t]
\centering
\begin{tabular}{@{}cccrrr@{}}
\toprule
\textbf{Model} & \textbf{FT size}                        & \textbf{PT/FT Language} & $|V_{PT}|$ & $|V_{FT}|$ & {\bf Overlap} \\ \midrule
\multirow{10}{*}{\begin{tabular}[c]{@{}c@{}}Obfuscated\\Pre-training\end{tabular}} &
  \multirow{2}{*}{\textbf{20k}} &
  src: nonsense-{\tt de/de} &
  1,289,374 &
  77,284 &
  119 \\
                 &                                & trg: nonsense-{\tt en/en}   & 680,221   & 56,339    & 15       \\ \cmidrule(l){2-6} 
                 & \multirow{2}{*}{\textbf{50k}}  & src: nonsense-{\tt de/de}   & 1,289,374 & 148,282   & 215      \\
                 &                                & trg: nonsense-{\tt en/en}   & 680,221   & 102,900   & 33       \\ \cmidrule(l){2-6} 
                 & \multirow{2}{*}{\textbf{100k}} & src: nonsense-{\tt de/de}   & 1,289,374 & 241,617   & 270      \\
                 &                                & trg: nonsense-{\tt en/en}   & 680,221   & 163,105   & 50       \\ \cmidrule(l){2-6} 
                 & \multirow{2}{*}{\textbf{500k}} & src: nonsense-{\tt de/de}   & 1,289,374 & 729,937   & 651      \\
                 &                                & trg: nonsense-{\tt en/en}   & 680,221   & 466,678   & 164      \\ \cmidrule(l){2-6} 
                 & \multirow{2}{*}{\textbf{1m}}   & src: nonsense-{\tt de/de}   & 1,289,374 & 1,170,435 & 950      \\
                 &                                & trg: nonsense-{\tt en/en}   & 680,221   & 730,119   & 271      \\ \midrule 
\multirow{10}{*}{\begin{tabular}[c]{@{}c@{}}Regular\\Pre-training\end{tabular}} &
  \multirow{2}{*}{\textbf{20k}} &
  src: {\tt de/de} &
  1,861,801 &
  77,284 &
  65,006 \\
                 &                                & trg: {\tt en/en}            & 1137,015  & 56,339    & 49,295   \\ \cmidrule(l){2-6} 
                 & \multirow{2}{*}{\textbf{50k}}  & src: {\tt de/de}            & 1,861,801 & 148,282   & 117,827  \\
                 &                                & trg: {\tt en/en}            & 1,137,015 & 102,900   & 85,111   \\ \cmidrule(l){2-6} 
                 & \multirow{2}{*}{\textbf{100k}} & src: {\tt de/de}            & 1,861,801 & 241,617   & 180,708  \\
                 &                                & trg: {\tt en/en}            & 1,137,015 & 163,105   & 126,278  \\ \cmidrule(l){2-6} 
                 & \multirow{2}{*}{\textbf{500k}} & src: {\tt de/de}            & 1,861,801 & 729,937   & 435,333  \\
                 &                                & trg: {\tt en/en}            & 1,137,015 & 466,678   & 291,138  \\ \cmidrule(l){2-6} 
                 & \multirow{2}{*}{\textbf{1m}}   & src: {\tt de/de}            & 1,861,801 & 1,170     & 600,922  \\
                 &                                & trg: {\tt en/en}            & 1,137,015 & 730,119   & 394,598  \\ \bottomrule
\end{tabular}%
\caption{Tokenized pre-training (PT) and fine-tuning (FT) word piece counts and overlap statistics comparing obfuscated pre-training (upper part) vs. regular pre-training (lower-part) for German-to-English parallel data with various fine-tuning data set sizes.}
\label{tab: obfuscation_word_overlap_de-en}
\end{table*}

\begin{table*}[t]
\centering
\resizebox{0.9\textwidth}{!}{%
\begin{tabular}{@{}cccrrr@{}}
\toprule
\textbf{Model} &
  \textbf{Language Pair} &
  \textbf{PT/FT Language} &
  $|V_{PT}|$ & $|V_{FT}|$ & {\bf Overlap} \\ \midrule
\multirow{6}{*}{\begin{tabular}[c]{@{}c@{}}Obfuscated\\Pre-training\end{tabular}} &
  \multirow{2}{*}{\tt{id-en}} &
  src: nonsense-{\tt de/id} &
  1,289,374 &
  18,095 &
  112 \\
 &                                 & trg: nonsense-{\tt en/en} & 680,221            & 18,167 & 0     \\ \cmidrule(l){2-6} 
 & \multirow{2}{*}{\tt{my-en}} & src: nonsense-{\tt de/my} & 1,289,374          & 1,598  & 1     \\
 &                                 & trg: nonsense-{\tt en/en} & 680,221            & 18,514 & 0     \\ \cmidrule(l){2-6} 
 & \multirow{2}{*}{\tt{tr-en}} & src: nonsense-{\tt de/tr} & 1,289,374          & 24,616 & 270   \\
 &                                 & trg: nonsense-{\tt en/en} & 680,221            & 26,236 & 0     \\ \midrule
\multirow{6}{*}{\begin{tabular}[c]{@{}c@{}}Regular\\Pre-training\end{tabular}} &
  \multirow{2}{*}{\tt{id-en}} &
  src: {\tt de/id} &
  1,861,801 &
  18,095 &
  3,722 \\
 &                                 & trg: {\tt en/en}         & {1,137,015} & 26,236 & 6,483 \\ \cmidrule(l){2-6} 
 & \multirow{2}{*}{\tt{my-en}} & src: {\tt de/my}         & 1,861,801          & 1,598  & 97    \\
 &                                 & trg: {\tt en/en}         & {1,137,015} & 18,514 & 4,407 \\ \cmidrule(l){2-6} 
 & \multirow{2}{*}{\tt{tr-en}} & src: {\tt de/tr}         & 1,861,801          & 24,616 & 5,569 \\
 &                                 & trg: {\tt en/en}         & {1,137,015} & 26,236 & 6,483 \\ \bottomrule
\end{tabular}%
}
\caption{Tokenized pre-training (PT) and fine-tuning (FT) word piece counts and overlap statistics comparing obfuscated pre-training (upper part) vs. regular pre-training (lower-part) for additional language directions.}
\label{tab: obfuscation_word_overlap_other}
\end{table*}

\subsection{Synthetic Pre-Training: Additional Language Pairs}
\label{appendix:supplementary-pretraining-results}

\cref{table:synthetic-pretraining-additional-pairs} shows translation decoding results (spBLEU) for additional non-English-centric language pairs. We compare synthetic pre-training on permuted binary trees vs. fine-tuning from a randomly initialized model as a function of the fine-tuning set size. Cells marked `n/a' indicate there was insufficient parallel data to create a fine-tuning set of the specified size.

\begin{table*}[htbp]
\centering
\begin{tabular}{clrrrrrrrr}
\toprule
& & \multicolumn{4}{c}{\textbf{OPUS-Test}} & \multicolumn{4}{c}{\textbf{FLORES-devtest}} \\ \cmidrule(lr){3-6} \cmidrule(lr){7-10}
{\bf Language Pair} & {\bf Model}          &  10k &  25k &  50k & 100k &  10k &  25k &  50k & 100k \\ \midrule
\multirow{2}{*}{{\tt de-id}} & random-init & 5.6 & 6.6 & 10.1 & 16.0 & 1.8 & 4.2 & 7.1 & 12.5 \\
& pb-trees & 6.4 & 11.7 & 16.0 & 19.8 & 4.1 & 8.7 & 12.4 & 16.3 \\ \midrule
\multirow{2}{*}{{\tt de-my}} & random-init & 10.7 & 15.2 & 19.6 & 23.6 & 1.4 & 2.7 & 4.2 & 5.9 \\
& pb-trees & 12.3 & 18.3 & 24.2 & 28.3 & 2.1 & 4.2 & 6.2 & 7.8 \\ \midrule
\multirow{2}{*}{{\tt id-my}} & random-init & 11.8 & 16.3 & 18.9 & \multirow{2}{*}{n/a} &  1.5 &  2.5 &  3.4 & \multirow{2}{*}{n/a} \\
& pb-trees    & 11.8 & 17.0 & 20.2 &      &  1.6 &  3.4 &  5.0 &      \\ \midrule
\multirow{2}{*}{{\tt id-tl}} & random-init & 15.2 & 17.6 & 20.9 & 23.5 &  0.2 &  0.3 &  0.4 &  0.6 \\
& pb-trees    & 16.7 & 18.5 & 21.8 & 24.8 &  0.5 &  0.9 &  1.5 &  2.9 \\ \midrule
\multirow{2}{*}{{\tt id-tr}} & random-init &  4.1 &  6.2 &  8.0 & 11.5 &  0.9 &  1.7 &  3.0 &  5.7 \\
& pb-trees    &  4.5 &  8.1 & 12.3 & 16.3 &  1.1 &  3.5 &  6.8 & 10.5 \\ \midrule
\multirow{2}{*}{{\tt my-tl}} & random-init & 11.9 & 16.4 & 21.6 & \multirow{2}{*}{n/a} &  2.0 &  2.8 &  3.7 & \multirow{2}{*}{n/a} \\
& pb-trees    & 12.8 & 19.6 & 27.0 &      &  2.4 &  4.3 &  5.8 &      \\ \midrule
\multirow{2}{*}{{\tt my-tr}} & random-init &  5.1 &  6.5 &  8.0 &  7.7 &  0.2 &  0.4 &  0.3 &  0.3 \\
& pb-trees    &  5.7 &  8.1 & 11.4 & 14.7 &  0.2 &  0.5 &  1.2 &  1.8 \\ \midrule
\multirow{2}{*}{{\tt tl-tr}} & random-init &  2.2 &  3.1 &  3.8 &  5.0 &  0.3 &  0.7 &  1.1 &  1.8 \\
& pb-trees    &  2.0 &  3.5 &  4.9 &  4.9 &  0.4 &  1.0 &  2.1 &  2.1 \\ \bottomrule
\end{tabular}
\caption{Translation decoding results for additional non-English-centric pairs. We report spBLEU for synthetic pre-training with pb-trees vs. fine-tuning from random initialization as a function of the fine-tuning set size. }
\label{table:synthetic-pretraining-additional-pairs}
\end{table*}

\section{Implementation Details}
\label{appendix:implementation_details}

This section describes implementation details for facilitating the reproduction of our work.

\subsection{Model Architectures}
All translation models described in our experiments are based on the sequence-to-sequence transformer `base' architecture \citep{DBLP:journals/corr/VaswaniSPUJGKP17} as implemented in {\tt fairseq} \citep{ott-fairseq-paper}. The models have six encoder layers, six decoder layers, and eight attention heads. The word embedding size is 512, and the feed-forward layers have 2048 dimensions. All BLEU scores are computed using SacreBLEU \citep{post-2018-call} with {\tt sentencepiece} tokenization \citep{goyal2022flores}. Our SacreBLEU scoring signature\footnote{{\tt BLEU+case.mixed+numrefs.1+smooth.exp\\+tok.spm+version.1.5.1}} indicates that both source and reference are {\tt sentencepiece} tokenized prior to scoring.

\subsection{Hyper-Parameters and Training Configuration}
\cref{tab:experiment_parameters} shows the hyper-parameters and training settings used for our experiments. We found different warm-up schedules were appropriate for the pre-training and fine-tuning stages. We choose the best model during training by maximizing the tokenized BLEU score on the validation set. For both pre-training and fine-tuning, we allow training to continue until the BLEU score has fully converged.

\begin{table*}[htpb]
\centering
\begin{tabular}{@{}lc@{}}
\toprule
\multicolumn{2}{c}{\textbf{Training Settings}} \\ \midrule
Optimizer                     & Adam                            \\
Learning Rate                 & 5e-4                            \\
Weight Decay                  & 1e-4                            \\
Criterion                     & label\_smoothed\_cross\_entropy \\
Label Smoothing               & 0.1                             \\
Learning Rate Scheduler       & Inverse sqrt                    \\
Warmup Updates (Pre-Training) & 4000                            \\
Warmup-Updates (Fine-Tuning)  & 100                             \\
Max Token Number              & 2048                            \\
Decoding Strategy             & Beam Search                     \\
Beam size                     & 5                               \\
Max Length a                  & 1.2                             \\
Max Length b                  & 10                              \\ \bottomrule
\end{tabular}%
\caption{Summary of pre-training and fine-tuning parameters for our experiments.}
\label{tab:experiment_parameters}
\end{table*}

\end{document}

%% file: math_commands.tex
\usepackage{amsmath,amsfonts,bm}

\def\eqref#1{equation~\ref{#1}}

\def\1{\bm{1}}

\DeclareMathAlphabet{\mathsfit}{\encodingdefault}{\sfdefault}{m}{sl}
\SetMathAlphabet{\mathsfit}{bold}{\encodingdefault}{\sfdefault}{bx}{n}

